\documentclass{article}
\usepackage{ijcai21}

\usepackage{times}
\usepackage{soul}
\usepackage{url}
\usepackage[hidelinks]{hyperref}
\usepackage[utf8]{inputenc}
\usepackage[small]{caption}
\usepackage{graphicx}
\usepackage{amsmath}
\usepackage{amsthm}
\usepackage{booktabs}
\usepackage{algorithm}
\usepackage{algorithmic}
\usepackage{amssymb}
\usepackage{multirow}
\DeclareMathOperator{\E}{\mathbb{E}}

\title{AutoReCon: Neural Architecture Search-based Reconstruction for Data-free Compression}

\author{
Baozhou Zhu$^{1,2}$
\and
Peter Hofstee$^{1,3}$\and
Johan Peltenburg$^{1}$\And
Jinho Lee$^{4}$\And
Zaid Alars$^1$
\affiliations
$^1$Delft University of Technology, Delft, The Netherlands\\
$^2$National University of Defense Technology, Changsha, China\\
$^3$IBM Austin, Austin, TX, USA\\
$^4$Yonsei University, Seoul, Korea
\emails
\{B.Zhu-1, j.w.peltenburg\, z.al-ars\}@tudelft.nl,
hofstee@us.ibm.com,
leejinho@yonsei.ac.kr
}

\begin{document}

\maketitle

\begin{abstract}
Data-free compression raises a new challenge because the original training dataset for a pre-trained model to be compressed is not available due to privacy or transmission issues. Thus, a common approach is to compute a reconstructed training dataset before compression. The current reconstruction methods compute the reconstructed training dataset with a generator by exploiting information from the pre-trained model. However, current reconstruction methods focus on extracting more information from the pre-trained model but do not leverage network engineering. This work is the first to consider network engineering as an approach to design the reconstruction method. Specifically, we propose the AutoReCon method, which is a neural architecture search-based reconstruction method. In the proposed AutoReCon method, the generator architecture is designed automatically given the pre-trained model for reconstruction. Experimental results show that using generators discovered by the AutoRecon method always improve the performance of data-free compression.
\end{abstract}

\section{Introduction}

To be deployed on resources-constrained hardware for real-time applications, the efficiency of deep convolutional neural networks has been improved significantly by various model compression techniques \cite{he2020learning,howard2019searching,zhu2020towards,mirzadeh2020improved}. Without altering the model architecture, quantized neural networks \cite{zhu2020towards} use a low bit width representation instead of full-precision floating-point, saving expensive multiplications. Pruning \cite{he2020learning} is an approach to remove the weights or neurons based on certain criteria. In terms of efficient neural network architectures, the MobileNet \cite{howard2019searching}, ShuffleNet, and ESPNet \cite{mehta2019espnetv2} series make use of depthwise-separable convolution, grouped convolution with shuffle operation, and efficient spatial pyramid. The knowledge distillation paradigm \cite{mirzadeh2020improved} transfers the information from a pre-trained teacher network to a portable student network. 

Data-free compression \cite{chen2019data,cai2020zeroq} has been an active research area when the original training dataset for the given pre-trained model is unavailable because of privacy or storage concerns.  Given the pre-trained model to be compressed, it is an essential step to reconstruct the original training dataset by inverting representation. For example, accuracy degradation of ultra-low precision quantized models \cite{banner2018aciq,xu2020generative,nagel2019data} is unacceptable without fine-tuning on the reconstructed training dataset. The reconstruction method computes a reconstructed training dataset by leveraging some extra metadata \cite{lopes2017data} or by extracting some prior information \cite{choi2020data} from the pre-trained model. Instead of computing the reconstructed training dataset directly \cite{nayak2019zero,cai2020zeroq,lopes2017data}, recent reconstruction methods \cite{fang2019data,yoo2019knowledge,micaelli2019zero,xu2020generative,choi2020data,chen2019data} employ a generator to generate a reconstructed training dataset in an end-to-end manner and show better performance for data-free compression.

The quality of the reconstruction closely relates to the extracted information from the pre-trained model. When more information is exploited from the pre-trained model, data-free compression achieves better performance. Thus, the current reconstruction methods \cite{micaelli2019zero,xu2020generative,nayak2019zero,chen2019data,choi2020data,yoo2019knowledge} focus on exploiting as much prior information as possible from the pre-trained model. However, how the network engineering will contribute to the reconstruction method remains unknown. Thus, we consider network engineering of the reconstruction method for the first time in the literature. This work aims to seek an optimized generator architecture, with which data-free compression shows performance improvement. It is worth mentioning that network engineering of the reconstruction and exploiting more prior information from the pre-trained model are complementary rather than contradictory. Both are important and should be explored for improving data-free compression. The contribution of this paper is summarized as follows.
\begin{itemize}
\item To our best knowledge, we are the first work to consider network engineering of the reconstruction method.
\item We propose the AutoReCon method, which is a neural architecture search-based reconstruction method to optimize generator architecture for reconstruction.
\item Using the discovered generator, diverse experiments are conducted to demonstrate the effectiveness of the AutoReCon method for data-free compression.
\end{itemize}

\section{Related Work}

\subsection{Neural Architecture Search}

Neural architecture search has attracted a lot of attention since it can automatically search for an optimized architecture for a certain task and achieve remarkable performance \cite{pham2018efficient,liu2018darts,gao2020adversarialnas,zhu2020nasb}. The optimization algorithms of neural architecture search include reinforcement learning \cite{pham2018efficient}, evolutionary algorithm, random search \cite{chen2018searching}, and gradient-based algorithm \cite{liu2018darts}. There is a lot of work towards reducing the computational resources required by searching, including weight sharing \cite{pham2018efficient}, progressive search, one-short mechanism \cite{liu2018darts}, and using a proxy task. The performance of the discovered architecture by neural architecture search has surpassed human-designed architecture in many computer vision tasks, including classification \cite{liu2018darts} and image generation \cite{gao2020adversarialnas}.

\subsection{Data-free Model Compression}

Data-free compression covers data-free quantization and data-free knowledge distillation. Without a generator, the reconstructed training dataset is computed directly in \cite{lopes2017data,nayak2019zero,cai2020zeroq,nagel2019data,yin2020dreaming}. \cite{lopes2017data} present a method for data-free knowledge distillation, where the reconstructed training dataset is computed based on some extra recorded activations statistics from the pre-trained model. DeepInversion \cite{yin2020dreaming} introduces a feature map regularizer based on batch normalization information in the pre-trained model for data-free knowledge distillation. In data-free knowledge distillation \cite{nayak2019zero}, the class similarities are computed from the pre-trained model and the output space is modeled via  Dirichlet Sampling. \cite{cai2020zeroq} calculates the reconstructed training dataset to match the statistics of the batch normalization layers of the pre-trained model and introduces the Pareto frontier to enable mixed-precision quantization. \cite{nagel2019data} improves data-free quantization by equalizing the weight ranges and correcting the biased quantization error.

The performance of data-free compression can be improved by employing a generator for the reconstruction \cite{fang2019data,yoo2019knowledge,micaelli2019zero,xu2020generative,choi2020data,chen2019data}. \cite{chen2019data} proposes a framework for data-free knowledge distillation by exploiting generative adversarial networks, where the reconstructed training dataset derivated from the generator is expected to lead to maximum response on the discriminator of the pre-trained model. The KEGNET \cite{yoo2019knowledge} framework uses the generator and decoder networks to estimate the conditional distribution of the original training dataset for data-free knowledge distillation. In data-free knowledge distillation \cite{micaelli2019zero}, an adversarial generator is used to produce and search for the reconstructed training dataset on which the student poorly match the teacher. In this paper, we improve on the work of \cite{xu2020generative}, which proposes a knowledge matching generator to produce a reconstructed training dataset by exploiting classification boundary knowledge and distribution information from the pre-trained model.

\section{AutoReCon Method for Data-free Compression}

In this section, we define the reconstruction method for data-free compression. Then, we introduce our proposed AutoReCon method, a neural architecture search-based reconstruction method, and present its search space and search algorithm. Also, the training process of the AutoReCon method for data-free compression is described.

\begin{figure}[tb]
\centering
\includegraphics[width=0.40\textwidth]{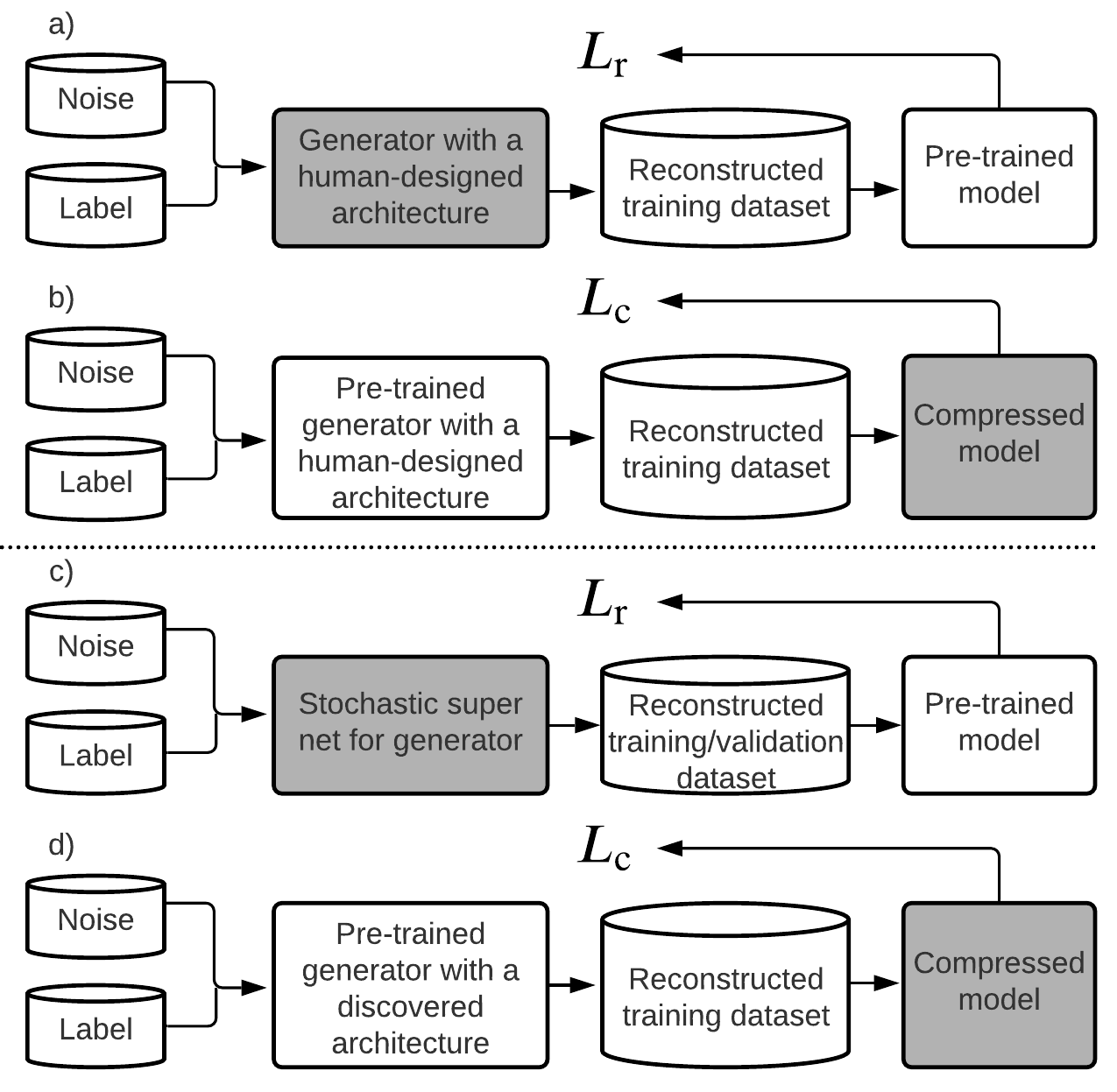}
\caption{The comparison between the current reconstruction method and the AutoReCon method for data-free compression. The goal of every subfigure is to update the models in gray color, given the pre-trained and fixed models in white color. \textbf{a)} an overview of the current reconstruction method to update the generator by minimizing the reconstruction loss $L_r$, where the generator has a human-designed architecture. \textbf{b)} an overview of the current reconstruction for data-free compression to update the compressed model by minimizing the compression loss $L_c$, after the generator with the human-designed architecture has been trained in subfigure \textbf{a)}. \textbf{c)} an overview of the AutoReCon method to update the generator by minimizing $L_r$, where there is a super net for the generator. \textbf{d)} an overview of the AutoReCon method for data-free compression to update the compressed model by minimizing $L_c$, after the generator with a discovered architecture has been trained in subfigure \textbf{c)}.}
\label{fig1}
\end{figure}

\subsection{Definition of Reconstruction Method}

The pre-trained model $M_p$ is obtained by training on the original training dataset $T_o=\left \{  x_o, y_o\right \}$. Given the pre-trained model $M_p$, we compute the reconstructed training dataset $T_r=\left \{  x_r, y_r\right \}$ with the reconstruction method $\Phi $, i.e., $T_r=\Phi(M_p)$. 

Considering the reconstruction method with a generator as shown in Figure~\ref{fig1}a), the pre-trained model $M_p$ is fixed while the weights of the generator are updated by minimizing the reconstruction loss $L_r$. The prior information extracted from the pre-trained model $M_p$ by the current methods is mainly the class boundary information and distribution information. If more prior information can be extracted from the pre-trained model, the reconstruction method can be easily adjusted by incorporating more loss terms to the reconstruction loss. Current reconstruction method $\Phi $ can be expressed as follows.
\begin{equation}
\begin{split}
&\operatorname*{min}_{W_g} L_{r}(W_g)= \operatorname*{min}_{W_g} \E_{y_o \sim {{P_{y_o}(y_o)}}, z \sim P_z(z)} \\ &[L_{class}(M_{p}(M_{g}(z|y_o);W_g), y_o) + L_{bns}(BN_r, BN_o)]  
\end{split}
\end{equation}
where $z$ and $W_g$ are the random noise input vector and weights of the generator, and $L_{class}(\cdot, \cdot)$ is the cross-entropy loss function. $L_{bns}(\cdot, \cdot)$ measures the distribution distance between the batch normalization statistics of the original training dataset $BN_o$ and the batch normalization statistics of the reconstructed training dataset $BN_r$. The formulations of $L_{class}$, $L_{bns}$, and $L_r$ are flexible to make the AutoReCon method general.

\begin{figure*}[tb]
\centering
\includegraphics[width=0.80\textwidth]{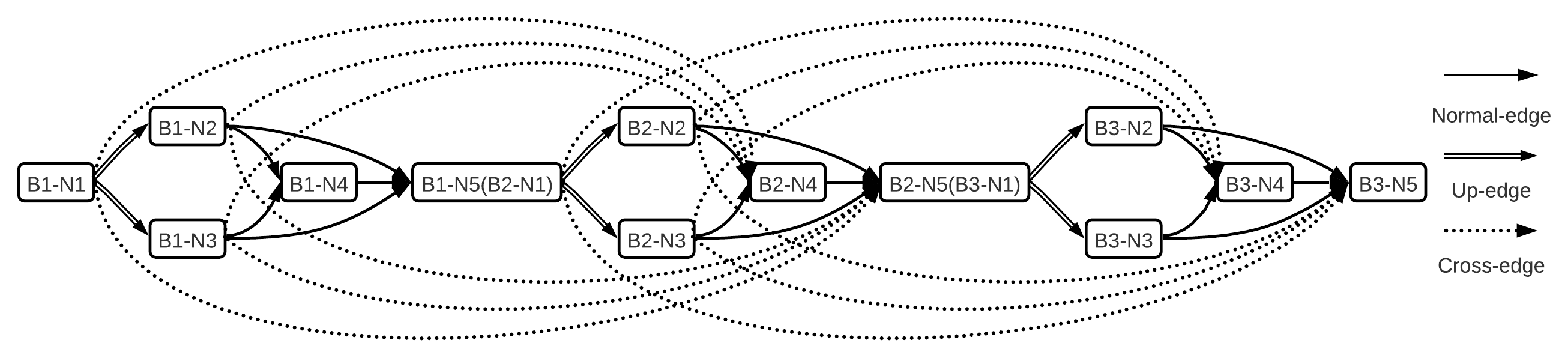}
\caption{The macro-architecture for the generator. The macro-architecture is a directed acyclic graph consisting of an ordered sequence of nodes. For example, the rectangle with the tag "B1-N1" represents the $1^\text{st}$ node of the $1^\text{st}$ convolutional block.  "B1-N5(B2-N1)" indicates the $5^{th}$ node of the $1^\text{st}$ convolutional block is the same as the $1^\text{st}$ node of the $2^\text{nd}$ convolutional block.}
\label{fig2}
\end{figure*}

\subsection{AutoReCon Method}

As shown in Figure~\ref{fig1}a) and c),  we present an overview of current reconstruction and the AutoReCon method. The current reconstruction method includes a pre-trained model $M_p$ and a generator $M_g$ with a human-designed architecture. In the AutoReCon method, we aim to search for a superior generator architecture automatically for reconstruction.

Regarding the reconstruction task, our training objective function is written as follows, where both weights $W_g$ and architecture $A_g$ of the generator can be updated by minimizing the reconstruction loss.
\begin{equation}
\begin{split}
&\operatorname*{min}_{A_g} L_r^{\text{val}}(A_g, W_g^*(A_g))  \\
&s.t.~W_g^*(A_g) = \operatorname*{argmin}_{W_g}~L_r^{\text{train}}(A_g, W_g) 
\label{equation2}
\end{split}
\end{equation}
where $L_r^{\text{train}}$ and $L_r^{\text{val}}$ refer to the reconstruction loss function on the reconstructed training dataset and the reconstructed validation dataset, respectively. $W_g^*(A_g)$ are the optimal weights of the generator given the generator architecture $A_g$. $A_g \in S$ and $S$ is the whole search space of the generator. 
\subsubsection{The Search Space} 

We construct a layer-wise search space with a fixed macro-architecture for the generator. The macro-architecture defines the type of the edge, the number of edges, the node connection, and the input/output dimension of each node. The macro-architecture is shown in Figure~\ref{fig2}, where there are three convolutional blocks and five nodes in every convolutional block. We denote the generator as $M_g(e_1, ..., e_i, ..., e_E)$, where $e_i$ represents the $i^{th}$ edge and $E$ is the number of edges. The nodes refer to the feature maps and we calculate them as the summation of the outputs of their previous connected edges. There are three types of edges: normal-edge, up-edge, and cross-edge. Normal-edge connects two nodes with the same dimension. Up-edge is used to increase the spatial resolution. Normal-edge and Up-edge are within a convolutional block. Cross-edge connects two adjacent convolutional blocks.

\begin{table}[tb]
\small{
\begin{center}
\begin{tabular}{ll}
\hline
Edge type  & Mixture of candidate operations  \\
\hline
\multirow{7}{*}{Normal-edge}  & Convolution 1 $\times$ 1, dilation=1  \\
~ &  Convolution 3 $\times$ 3, dilation=1 \\
~ &  Convolution 5 $\times$ 5, dilation=1 \\
~ &  Convolution 3 $\times$ 3, dilation=2 \\
~ &  Convolution 5 $\times$ 5, dilation=2 \\
~& Identity \\
~& None \\
\hline
\multirow{2}{*}{Up-edge}  & Nearest Neighbor Interpolation \\
~ & Bilinear Interpolation \\
\hline
\multirow{3}{*}{Cross-edge}  & Nearest Neighbor Interpolation \\
 & Bilinear Interpolation \\
  & None   \\
\hline
\end{tabular}
\end{center}
\caption{For different types of edges, there are different mixtures of candidate operations.}
\label{table1}
}
\end{table}

To construct a layer-wise search space for the generator, we set each edge as a mixture of candidate operations, which has several parallel operations instead of one specific operation. Thus, the over-parameterized generator is expressed as $M_g(e_1=C_1, ..., e_i=C_i, ..., e_E=C_E)$ and $C_i$ is the mixture of candidate operations for the edge $e_i$. As shown in Table~\ref{table1}, different types of edges use different mixtures of candidate operations. Taking the edge $C_i$ as an example, we compute its output by summing the outputs of the mixture of candidate operations as follows. 
\begin{equation}
X_{out}^i= C_i(X_{in}^i) = \sum_{j=1}^{F}O_j^{i}(X_{in}^i) 
\label{equation3}
\end{equation}
where $X_{in}^i$ and $X_{out}^i$ are the input and output of the $i^{th}$ edge. $O_j^{i}$ denotes the $j^{th}$ candidate operation of the $i^{th}$ edge and $j=1, ..., F$. $F$ is the number of candidate operations for an edge.

\subsubsection{The Search Algorithm} 

The search algorithm represents the search space as a stochastic super net $M_s$. In the stochastic super net $M_s$, $O_j^{i}$ is associated with an architecture parameter $\alpha  _j^i$. To derive a generator $A_g$ from the stochastic super net $M_s$, the candidate operation $O_j^{i}$ is sampled with the probability $p_j^{i}$, which is computed as follows.
\begin{equation}
p_j^{i}(O_j^{i};\alpha^i) = \text{softmax}(\alpha^i)=\frac{exp(\alpha _j^i)}{\sum_{j=1}^{F}exp(\alpha _j^i)} 
\end{equation}
Since sampling from the mixture of candidate operations for each edge is independent, the probability of sampling a generator architecture $A_g $ can be described as follows.
\begin{equation}
P(A_g ;\alpha_g)=\prod_{i=1}^{E}p_j^{i}(O_j^{i};\alpha^i)
\end{equation}
In this case, we can approximate the problem of finding an optimized discrete generator architecture by finding optimized sampling probabilities. The training objective function of the AutoReCon method is re-written from Equation~\ref{equation2} as follows.
\begin{equation}
\begin{split}
&\operatorname*{min}_{a_g} 
\E_{a_g \sim P_{a_g}(a_g)}   
[L_r^{\text{val}}(a_g, W_g^*(a_g))]  \\
&s.t.~W_g^*(a_g) = \operatorname*{argmin}_{W_g}~L_r^{\text{train}}(a_g, W_g) 
\end{split}
\end{equation}
To make the reconstruction loss differentiable to the sampling probabilities, we compute continuous variables $m_j^{i}$ by the Gumbel Softmax function as an alternative as follows.
\begin{equation}
m_j^{i} = \text{GumbelSoftmax}(p^i)=\frac{exp\left [(p _j^i + g_j^i)/\tau  \right ]}{\sum_{j=1}^{F}exp\left [(p _j^i + g_j^i)/\tau  \right ]} 
\end{equation}
where $g_j^i$ is the noise sampled from the Gumbel distribution ($0, 1$) and $\tau $ is a temperature parameter to control the sampling operation. Then, the continuous variables $m_j^{i}$ are directly differentiable with respect to the sampling probabilities. Thus, the computation of the edge $C_i$ in Equation~\ref{equation3} can be expressed as follows. 
\begin{equation}
X_{out}^i= C_i(X_{in}^i) = \sum_{j=1}^{F}m_j^{i}O_j^{i}(X_{in}^i)
\end{equation}

 \begin{algorithm}[tb]
 \small{
 \caption{The AutoReConmethod for data-free compression}
 \label{algorithm1}
 \begin{algorithmic}[1]
 \renewcommand{\algorithmicrequire}{\textbf{Input:}}
 \renewcommand{\algorithmicensure}{\textbf{Output:}}
 \REQUIRE Pre-trained model $M_p$.
 \ENSURE Discovered generator $M_g$, compressed model $M_c$. \\
 \textbf{Stage 1:} Searching for generator architecture.\\
 \FOR {$epoch = 1$ to $L_1$}
  \FOR {$batch = 1$ to $T_1$}
   \STATE 
    Obtain random noise $z  \sim  N (0, 1)$ and label $y_o$. \\
    Generate reconstructed training dataset $T_r$ with stochastic super net $M_s$. \\
    Update weights of stochastic super net by minimizing reconstruction loss $L_r$. \\
  \ENDFOR
 \FOR {$batch = 1$ to $V_1$}
   \STATE 
    Obtain random noise $z  \sim  N (0, 1)$ and label $y_o$. \\
    Generate reconstructed validation dataset $V_r$ with stochastic super net $M_s$. \\
    Update architecture parameters of stochastic super net by minimizing reconstruction loss $L_r$. \\
  \ENDFOR
  \ENDFOR \\
   \textbf{Stage 2:} Compression with discovered generator.\\
 \FOR {$epoch = 1$ to $L_2$}
  \FOR {$batch = 1$ to $T_2$}
   \STATE 
    Obtain random noise $z  \sim  N (0, 1)$ and label $y_o$. \\
    Generate reconstructed training dataset $T_r$ with the discovered generator $M_g$. \\
    Update weights of compressed model $M_c$ by minimizing compression loss $L_c$. \\
  \ENDFOR
  \ENDFOR
 \end{algorithmic}
 }
 \end{algorithm}

\subsection{Training Process}

Using the AutoReCon method, the training process for data-free compression is illustrated as shown in Algorithm~\ref{algorithm1}. The first stage of the training process is to search for generator architecture with our AutoReCon method, as shown in Figure~\ref{fig1}c). The goal of the first stage is to seek an optimized generator architecture from the stochastic super net. The second stage of the training process is to compress the pre-trained model $M_p$ with the discovered generator $M_g$. The compression loss $L_c$ can be introduced from quantization and/or knowledge distillation. Compared with the current reconstruction methods, our AutoReCon method considers network engineering and search for an optimized generator architecture for reconstruction.

\begin{table*}[tb]
\small{
\begin{center}
\begin{tabular}{llllll}
\toprule
Method & Pre-trained model &
Generator
& Quantization
& Top-1(CIFAR-100) & Top-1(ImageNet)
  \\
\hline
- & ResNet18 & - & - & $78.83\%$ & - \\
- & ResNet18 & -  & - & -        & $ 71.47\%$ \\
GDFQ & ResNet18 & Human-designed & w6a6  & $78.00\%$  & $70.10\%$ \\
GDFQ & ResNet18 & Human-designed & w5a5  & $75.93\%$  & $68.38\%$ \\
GDFQ & ResNet18 & Human-designed & w4a4  & $60.23\%$  & $60.70\%$ \\
GDFQ & ResNet18 & Human-designed & w3a3  & $28.71\%$  & $20.69\%$ \\
Ours & ResNet18 & Discovered by AutoReCon & w6a6  & $78.52\%$($+0.52\%$)   & $70.61\%$($+0.51\%$) \\
Ours & ResNet18 & Discovered by AutoReCon & w5a5  & $77.22\%$($+1.29\%$)   & $68.88\%$($+0.50\%$) \\
Ours & ResNet18 & Discovered by AutoReCon & w4a4  & $71.02\%$($+10.79\%$)  & $61.32\%$($+0.62\%$)  \\
Ours & ResNet18 & Discovered by AutoReCon & w3a3  & $46.44\%$($+17.73\%$)  & $23.37\%$($+2.68\%$) \\
\hline
- & MobileNetV2 & - & - & $70.72\%$ & - \\
- & MobileNetV2 & - & - & -         & $73.03\%$ \\
GDFQ & MobileNetV2 & Human-designed & w6a6  & $69.59\%$   & $71.18\%$ \\
GDFQ & MobileNetV2 & Human-designed & w5a5  & $65.27\%$   & $67.81\%$ \\
GDFQ & MobileNetV2 & Human-designed & w4a4  & $53.91\%$   & $59.80\%$ \\
GDFQ & MobileNetV2 & Human-designed & w3a3  & $8.50\%$    & $2.31\%$ \\
Ours & MobileNetV2 & Discovered by AutoReCon & w6a6  & $70.57\%$($+0.98\%$)  & $71.53\%$($+0.33\%$) \\
Ours & MobileNetV2 & Discovered by AutoReCon & w5a5  & $67.95\%$($+2.68\%$)  & $68.40\%$($+0.59\%$) \\
Ours & MobileNetV2 & Discovered by AutoReCon & w4a4  & $58.42\%$($+4.51\%$)  & $60.13\%$($+0.33\%$) \\
Ours & MobileNetV2 & Discovered by AutoReCon & w3a3  & $10.21\%$($+1.71\%$)  & $14.30\%$($+11.99\%$) \\
\hline
- & ResNet50 & - & - & $79.36\%$ & - \\
- & ResNet50 & - & - & - & $77.72\%$ \\
GDFQ & ResNet50 & Human-designed & w6a6  & $78.79\%$    & $76.40\%$ \\
GDFQ & ResNet50 & Human-designed & w5a5  & $76.17\%$    & $70.79\%$ \\
GDFQ & ResNet50 & Human-designed & w4a4  & $61.44\%$    & $55.94\%$ \\
GDFQ & ResNet50 & Human-designed & w3a3  & $26.51\%$    & $1.20\%$ \\
Ours & ResNet50 & Discovered by AutoReCon & w6a6  & $79.12\%$($+0.33\%$)  & $76.76\%$($+0.36\%$) \\
Ours & ResNet50 & Discovered by AutoReCon & w5a5  & $77.06\%$($+0.89\%$)  & $74.13\%$($+3.34\%$) \\
Ours & ResNet50 & Discovered by AutoReCon & w4a4  & $68.20\%$($+6.76\%$)  & $64.37\%$($+8.43\%$) \\
Ours & ResNet50 & Discovered by AutoReCon & w3a3  & $36.17\%$($+9.66\%$) &  $1.63\%$($+0.43\%$) \\
\bottomrule
\end{tabular}
\end{center}
\caption{Experimental results of data-free compression on CIFAR-100 and ImageNet classification. $w4a4$ means that the weights and activations are quantized to $4$-bit precision. Both our data-free compression method and the GDFQ adopt knowledge distillation for the output layer. In each block, the first row presents the accuracy of the full-precision pre-trained model on CIFAR-100.  The second row shows the accuracy of the full-precision pre-trained model on ImageNet.}
\label{table2}
}
\end{table*}

\section{Experiments}

\subsection{Implementation Details}

Our interest is to show the performance improvement of data-free compression, which is brought by the AutoReCon method. We adopt the GDFQ data-free compression method \cite{xu2020generative} as a baseline for the following three reasons. First, it exploits both class boundary information and distribution information from the pre-trained model $M_p$, compared to other methods that use only one type of information \cite{cai2020zeroq,yoo2019knowledge,chen2019data,nayak2019zero}. Second, it includes both data-free quantization and data-free knowledge distillation, where knowledge distillation is applied for the output layer (i.e., knowledge distillation is not applied for the intermediate layers). Third, it achieves state-of-the-art performance. We use the same experimental settings as the GDFQ method to observe the influence of the generator architecture. In the GDFQ method, the human-designed generator architecture for both CIFAR-100 and ImageNet classification follows ACGAN. Besides, the human-designed generator for ImageNet classification adopts the categorical conditional batch normalization layer to fuse label information following SN-GAN.

\subsection{Results on Image Classification}

\subsubsection{Results on ImageNet Classification} 

As shown in Table~\ref{table2}, we report the experimental results of data-free compression on the ImageNet classification dataset. Replacing the human-designed generator with the generator discovered by our AutoReCon method, the accuracy of the GDFQ method increases consistently using different pre-trained models and low-bit width quantization. Using ResNet18 as the pre-trained model and $3$-bit width quantization, the Top-1 accuracy of the GDFQ method can increase by $2.68\%$ when using the generator discovered by the AutoReCon method. The Top-1 accuracy of the GDFQ method increases by $11.99\%$ using MobileNetV2 as the pre-trained model, $3$-bit width quantization, and the generator discovered by the AutoReCon method. Using ResNet50 as the pre-trained model and $5$-bit width quantization, the Top-1 accuracy of our data-free compression with an optimized generator surpasses the GDFQ method by $8.43\%$. In addition, the optimized generator needs almost the same parameters and fewer flops compared with a human-designed generator.

\subsubsection{Results on CIFAR-100 Classification} 

As shown in Table~\ref{table2}, we report the experimental results of data-free compression on the CIFAR-100 classification dataset. Using various pre-trained models and low-bit width quantization, our data-free compression with an optimized generator architecture achieves better accuracy than the GDFQ method with a human-designed generator. Using ResNet18 as the pre-trained model and $3$-bit width quantization, the Top-1 accuracy of the GDFQ method will improve by $17.73\%$ if the human-designed generator is replaced with the generator discovered by the AutoReCon method. Using MobileNetV2 and $5$-bit width quantization, the Top-1 accuracy of our data-free compression shows an improvement of $4.51\%$ compared with the GDFQ method. The Top-1 accuracy improvement becomes $9.66\%$ using ResNet50 as the pre-trained model and $4$-bit width quantization.

\subsection{Ablation Study}

\begin{table}[tb]
\small{
\begin{center}
\begin{tabular}{llll}
\hline
Method  &
Scale
& Top-1 & Top-5
  \\
\hline
GDFQ  & $s=4$   & $64.87\%$ & $86.76\%$ \\
GDFQ  & $s=3$   & $65.04\%$ & $86.93\%$ \\
GDFQ  & $s=2$   & $65.22\%$ & $87.19\%$ \\
GDFQ  & $s=1$   & $65.27\%$ & $87.30\%$ \\
GDFQ  & $s=0.5$   & $63.72\%$ & $86.21\%$ \\
Ours  & $s=4$   & $68.78\%$($+3.91\%$) & $88.62\%$ \\
Ours  & $s=3$   & $68.09\%$($+3.05\%$) & $89.01\%$ \\
Ours  & $s=2$   & $67.95\%$($+2.73\%$) & $88.76\%$ \\
Ours  & $s=1$   & $67.58\%$($+2.31\%$) & $88.42\%$ \\ 
Ours  & $s=0.5$   & $66.30\%$($+2.58\%$) & $88.09\%$ \\ 
\hline
\end{tabular}
\end{center}
\caption{Experimental results of data-free compression on CIFAR-100 classification. The GDFQ method uses a human-designed generator. Our data-free compression uses the generator discovered by the AutoRe method.}
\label{table6}
}
\end{table}

\subsubsection{Scalability of Discovered Generator Architectures} 

We explore the scalability of the discovered generator architecture for data-free compression on the CIFAR-100 classification dataset. We scale the base channels by a factor from $s=0.5$ to $s=4$ for the discovered generator and the human-designed generator. The data-free compression results using MobileNetV2 as the pre-trained model, $5$-bit width quantization, and knowledge distillation applied for the output layer are shown in Table~\ref{table6}. Without modifying the optimized generator architecture, the performance of our data-free compression keeps increasing and is always better than the GDFQ method when scaling the base channels by the factor from $s=0.5$ to $4.0$. The accuracy of the GDFQ method decreases when we scale the base channels for the human-designed generator. Thus, we conclude that our searched generator architecture has superior scalability compared to the human-designed generator for data-free compression.

\begin{table}[tb]
\small{
\begin{center}
\begin{tabular}{lll}
\hline
Method  &
Generator
& Top-1 
  \\
\hline
-  & -   & $77.50\%$  \\
DAFL  & Human-designed   & $61.40\%$  \\
DFAD  & Human-designed   & $67.70\%$  \\
Ours  & Discovered by AutoReCon   & $69.98\%$($+2.28\%$)  \\
\hline
\end{tabular}
\end{center}
\caption{Experimental results of data-free compression on CIFAR-100 classification. The first row is the accuracy of the pre-trained teacher model.}
\label{table7}
}
\end{table}

\subsubsection{Generalization of AutoReCon Method} 

Except for the GDFQ method, we use the GFAD\cite{fang2019data} method as a baseline to show the generalization of our AutoReCon method. The generation loss in the GFAD method is replaced with the reconstruction loss of Equation~6, which enables the exploration of generator architecture. We use ResNet34 as the pre-trained teacher model and ResNet18 as the student model. The experimental results of data-free knowledge distillation on CIFAR-100 is shown in Table~\ref{table7}. With a human-designed generator, the GFAD method achieves better accuracy than the DAFL\cite{chen2019data} method. The Top-1 accuracy of our data-free knowledge distillation with a discovered generator is $2.28\%$ better than the baseline of the GFAD method with a human-designed generator.

\begin{table}[tb]
\small{
\begin{center}
\begin{tabular}{llll}
\hline
Method & Pre-trained model 
& Quantization
& Top-1   \\
\hline
- & ResNet18 & -  & $ 71.47\%$  \\
DFQ & ResNet18  & w4a4  & $0.10\%$  \\
ZeroQ & ResNet18  & w4a4  & $26.04\%$  \\
DFC & ResNet18  & w4a4  & $55.49\%$  \\
GDFQ & ResNet18  & w4a4  & $60.70\%$  \\
Ours & ResNet18  & w4a4  & $61.60\%$  \\
\hline
- & MobileNetV2 & -  & $73.03\%$  \\
DFQ & MobileNetV2  & w4a4 &  $0.11\%$  \\
ZeroQ & MobileNetV2  & w4a4 &  $3.31\%$  \\
GDFQ & MobileNetV2  & w4a4 &  $59.80\%$  \\
Ours & MobileNetV2  & w4a4 & $60.02\%$ \\
\hline
- & ResNet50  & - & $77.72\%$ \\
ZeroQ & ResNet50 & w4a4 & $0.12\%$ \\ 
GDFQ & ResNet50  & w4a4 & $55.94\%$ \\
Ours & ResNet50 & w4a4  & $57.49\%$ \\
\hline
\end{tabular}
\end{center}
\caption{Comparison of different data-free compression methods on ImageNet classification. $w4a4$ means that the weights and activations are quantized to $4$-bit precision. The first row of each block is the accuracy of the full-precision pre-trained model.}
\label{table5}
}
\end{table}

\subsection{Comparison with State-of-the-art Methods}

On the ImageNet classification dataset, we present the results of additional data-free compression methods as shown in Table~\ref{table5}. The comparison is mainly for data-free quantization except that the GDFQ and our methods apply knowledge distillation on the output layer. None of the compared methods apply knowledge distillation on the intermediate layers. The results of DFQ \cite{nagel2019data} and ZeroQ \cite{cai2020zeroq} are cited from the GDFQ paper and have a rather low accuracy for ultra-low precision data-free quantization. The DFC \cite{haroush2020knowledge} method achieves a moderate accuracy with a combination of BN-Statistics and Inception schemes. Our method achieves better accuracy compared to the GDFQ method since the AutoReCon method discovers an optimized generator architecture for reconstruction. 

\section{Conclusion}

In this paper, we present the AutoReCon method, which is the first work to consider network engineering of the reconstruction method to improve the performance of data-free compression. In particular, our AutoReCon method can search for an optimized generator architecture from a stochastic super net with gradient-based neural architecture search for reconstruction. When we plug our discovered generator to replace the human-designed generator, our data-free compression benefits from the optimization of the generator architecture and achieves better accuracy. Specifically, using ResNet50 as the pre-trained model and $5$-bit width quantization, the Top-1 accuracy of our data-free compression on ImageNet with an optimized generator surpasses the GDFQ method by $8.43\%$. The Top-1 accuracy of the DFAD method on CIFAR-100 increases by $2.28\%$ using ResNet34 as the pre-trained teacher model, ResNet18 as the student model, and the generator discovered by the AutoReCon method.

\bibliographystyle{named}
\bibliography{ijcai21}

\end{document}